# Learning Finite-State Controllers for Partially Observable Environments

**Nicolas Meuleau, Leonid Peshkin, Kee-Eung Kim** and **Leslie Pack Kaelbling**
Computer Science Department, Box 1910, Brown University, Providence, RI 02912-1910
{nm, ldp, kek, lpk}@cs.brown.edu

## Abstract

Reactive (memoryless) policies are sufficient in completely observable Markov decision processes (MDPs), but some kind of memory is usually necessary for optimal control of a partially observable MDP. Policies with finite memory can be represented as finite-state automata. In this paper, we extend Baird and Moore's VAPS algorithm to the problem of learning general finite-state automata. Because it performs stochastic gradient descent, this algorithm can be shown to converge to a locally optimal finite-state controller. We provide the details of the algorithm and then consider the question of under what conditions stochastic gradient descent will outperform exact gradient descent. We conclude with empirical results comparing the performance of stochastic and exact gradient descent, and showing the ability of our algorithm to extract the useful information contained in the sequence of past observations to compensate for the lack of observability at each time-step.

## 1 INTRODUCTION

Learning an optimal policy in a large partially observable environment is a recurrent problem in many application domains of AI. However, there is no known technique that scales up well to increasing size *and difficulty* of the problem. This situation is due in part to the fact that planning in partially observable environments is itself a difficult task, hence learning to plan cannot be much easier. The partially observable Markov decision process (POMDP) provides a formal framework for studying these problems [1, 27, 28, 7, 11, 6, 16]. The difficulty of planning in partially observable environments is illustrated by the fact that the optimal policy of a POMDP may use the complete previous history of the system (i.e., the whole sequence of observations, actions and rewards since time 0) to determine the next action to perform. Therefore, we need an infinite memory if we want to act optimally over an infinite horizon.

A general way to represent policies is in the form of state-automata, or as we will call them, policy graphs. Every policy has a representation in the form of a (possibly infinite) policy graph. *A priori*, the optimal solution of a POMDP may well be an infinite policy graph. However, because of evident computational limits, we may reduce the search to policies representable as *finite* policy graphs. Many existing algorithms for learning to plan in POMDPs rely on a similar assumption. For instance, some researchers [14, 3, 32] try to learn memoryless (or reactive) policies, McCallum's learning algorithm [19, 20] uses a finite-horizon memory, Wiering and Schmidhuber's HQL [31] learns finite sequences of reactive policies using an implicit memory of some of the previous observations, and Peshkin et al. [23] look for optimal finite-external-memory policies. All these finite-memory architectures correspond to finite policy graphs with a particular structure in each case (i.e. *not* every node-transition and choice of action is possible in the graph). [1]

Most previous examples of search in the finite policy-graph-space [25, 8, 9, 10] use the criterion of $\epsilon$-optimality: they search for a finite graph whose value is less than $\epsilon$ *from the value of the optimal—Bayesian—solution*. Therefore, they need to work explicitly in the continuous space of belief functions, which is a cumbersome and sometimes intractable process. Another approach uses EM to find a finite controller that is optimal over a finite horizon [12].

In a companion paper [22], we proposed to solve problems with a very large state-space by fixing the size of the policy graph and trying to find the best graph of this size. We may then hope to find a graph-size that realizes a good compromise between the quality of the solution and the time required for finding it. This approach allows by-passing the belief-state space, and performing all the computation in a discrete setting, like in completely observable Markov decision processes (MDPs) [13, 24].[2] However, the algorithms do not provide any evaluation of the quality of the solution produced *relative to the optimal performance.*

[1] Note that, even though they can remember only a finite number of events, general (unconstrained) finite policy graphs can remember events arbitrarily far in the past.

As we showed in the companion paper [22], finding the best finite policy graph of a given size is NP-hard. However, some classical optimization techniques such as branch-and-bound search and gradient descent can be accelerated using previous knowledge about the structure of the problem at hand and its optimal solution. Despite this leverage, these techniques do not escape enumerating the set of all states of the POMDP at least once per iteration (they are at least in $O(|S|)$, where $S$ is the state-space of the POMDP). Hence, they cannot be applied to problems with a very large number of states, such as combinatorial problems where the state of the process is a vector of several state-features, and, therefore, the number of states is exponential in the number of features. Moreover, they require a complete initial knowledge of the parameters of the POMDP, i.e., they cannot be used to learn a policy without first learning a model of the environment.

Direct (model-free) learning of a policy during a (possibly simulated) interaction with the process is becoming a classical technique for planning in very large state-spaces [4, 17, 30]. The idea is to perform stochastic gradient descent by sampling state transitions and rewards during the experience. Because we sample only possible (and even reasonably probable) trajectories, the algorithm may be much more efficient than an exact method that enumerates every trajectory, including impossible and very low probability ones. This is the principle at the basis of most successful application of reinforcement learning (RL) to real world problems.

In this paper, we propose a model-free algorithm for learning general finite policy graphs of a given size. This algorithm can be used to learn finite-memory policies in some environments with a large number of states. As it is performing stochastic gradient descent in the parameters of the policy graph, it is ensured to converge to a local optimum. It is basically an extension of Baird and Moore's VAPS algorithm [3] for learning simple reactive policies. This constitutes a significant improvement to the original VAPS, since the restriction to reactive policies is a severe handicap in most partially observable domains. [3]

The paper is organized as follows. First, we give a quick introduction to POMDPs and policy graphs. Second, we develop the formalism of Baird and Moore's VAPS algorithm in the general framework of finite policy graphs. This represents the main contribution of the paper. Then we discuss the conditions under which stochastic gradient descent can outperform exact gradient descent (which is possible only when the problem is known in advance). Finally, we use the pole-balancing problem to show that our algorithm can solve difficult real-world problems with limited observability of the state of the system.

[2]However the optimality criterion used is the same as in the Bayesian approach, i.e., the expected discounted cumulative rewards, the expectation being relative to the prior belief on the states

[3]Singh et al. [26] showed that stochastic reactive policies can perform arbitrarily better than deterministic ones. However, it is also proven that the best stochastic reactive policy can be arbitrarily worse that the optimal memory-based policy.

## 2 POMDPs AND FINITE POLICY GRAPH

### 2.1 POMDPs

A partially observable Markov decision process (POMDP) is defined as a tuple $\langle S, O, A, B, T, R \rangle$ where:

- $S$ is the (finite) set of states;
- $O$ is the (finite) set of observations;
- $A$ is the (finite) set of actions;
- $B(s, o) = \Pr(o^t = o \mid s^t = s)$ for all $t$;
- $T(s, a, s') = \Pr(s^{t+1} = s' \mid s^t = s, a^t = a)$ for all $t$;
- $r^t = R(s, a, s')$ if $s^t = s$, $a^t = a$ and $s^{t+1} = s'$, for all $t$.

The underlying Markov decision process (MDP) $\langle S, A, T, R \rangle$ is optimized in the following way [13, 24]: given an initial state $s^0$, the aim is to maximize the expected discounted cumulative reward

$$\mathrm{E}\left(\sum_{t=0}^{\infty} \gamma^t r^t \mid s^0\right),$$

where $\gamma \in [0, 1)$ is the discount factor. The optimal solution is a mapping $\mu^* : S \to A$. It is a remarkable property of MDPs that there exists an optimal policy that always executes the same action in the same state. Unfortunately, this policy can not be used in the partially observable framework, because of the residual uncertainty on the current state of the process.

In a POMDP, a policy is a rule specifying the action to perform at each time step as a function of the whole previous history, i.e., the complete sequence of observation-action pairs since time 0. A particular kind of policy, the so-called reactive policies (RPs), condition the choice of the next action only on the last observation. Thus, they can be represented as mappings $\mu : O \to A$. Given a probability distribution $\pi^0$ over the starting state, each policy $\mu$ (reactive or not) realizes an expected cumulative reward:

$$\mathrm{E}\left(\sum_{t=0}^{\infty}\gamma^t r^t \mid \pi^0, \mu\right). \tag{1}$$

The classical—Bayesian—approach allows us to determine the policy that maximizes this value. It is based on updating the state distribution (or belief) at each time step, depending on the most recent observations [7, 11, 6, 16]. The problem is reformulated as a new MDP using belief-states instead of the original states. Generally, the optimal solution is not a reactive policy. It is a sophisticated behavior, with optimal balance between exploration and exploitation. Unfortunately, the Bayesian calculation is highly intractable as it searches the continuous space of beliefs and considers every possible sequence of observations.

## 2.2 FINITE POLICY GRAPHS

A policy graph for a given POMDP is a graph where the nodes are labeled with actions $a \in A$, the arcs are labeled with observations $o \in O$, and there is one and only one arc emanating from each node for each possible observation. When the system is in a certain node, it executes the action associated with this node. This implies a state transition in the POMDP and eventually a new observation (which depends on the arrival state of the underlying MDP). This observation itself conditions a transition in the policy graph to the destination node of the arc associated with the new observation. We are interested in stochastic policy graphs where action-choices and node-transitions are probabilistic. We will use the following notation:

- $N$ is the set of nodes of the graph,
- $n^t \in N$ is the current node at time $t$,
- $\psi(n, a)$ is the probability of choosing action $a$ in node $n \in N$:

  $$\psi(n,a) \stackrel{\text{def}}{=} \Pr(a^t = a \mid n^t = n), \text{ for all } t,$$

- $\eta(n, o, n')$ is the probability of moving from node $n \in N$ to node $n' \in N$, after observation $o \in O$:

  $$\eta(n,o,n') \stackrel{\text{def}}{=} \Pr(n^{t+1} = n' \mid n^t = n \wedge o^{t+1} = o),$$

  for all $t$,

- $\eta^0$ is the probability distribution of the initial node $n^0$ conditioned on the first observation $o^0$:

  $$\eta^0(o,n) \stackrel{\text{def}}{=} \Pr(n^0 = n \mid o^0 = o).$$

Figure 1 illustrates the functioning of policy graphs in POMDPs.

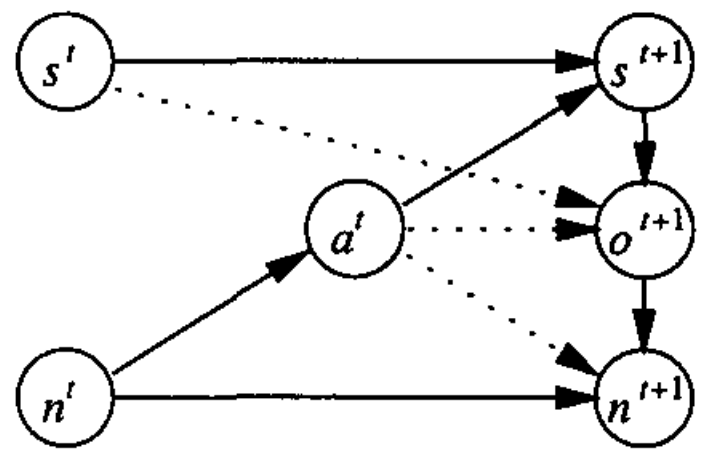

Figure 1: Influence diagram illustrating functioning of policy graphs in POMDPs. Dotted arrows represent dependencies that we did not take into account here, but that are sometimes represented in other formulations.

Every policy has a representation as a possibly infinite policy graph. A policy that chooses a different action for each possible previous history will be represented by an infinite tree with a branch for each possible history. Reactive policies correspond to a special kind of finite policy graph with as many nodes as there are observations in the POMDP, and whose structure is fixed. Other finite-memory architectures such as HQL's finite RP-sequences and finite external memory policies also correspond to finite policy graph with special structural constraints. [4]

## 2.3 FINDING AN OPTIMAL POLICY GRAPH

The problem of finding the optimal policy graph of a given size is studied in a companion paper [22]. The principle of this study is to exploit the Markov property of the association between POMDPs and finite policy graphs. It ends up with the proposition of algorithms that scale up relatively well with respect to the size of the problem, but that are more sensitive to the size of the policy graph.

Although these methods enable the solution of problems with up to 1000 states in a reasonable time, this approach is fundamentally limited by the necessity to enumerate the complete set of states of the POMDP, at least once at each time step. Thus, they will fail to solve problems with exponentially many states, such as the huge combinatorial problems often met in the real world. Moreover, these algorithms are basically planning algorithms, i.e., they require complete and accurate preliminary knowledge of the POMDP parameters, and they cannot be used to learn the policy on-line.

A possible solution to overcome the curse of dimensionality of the state space consists of having a direct (model-free) reinforcement learning (RL) algorithm learn a policy during a (possibly simulated) interaction with the process [4, 17, 30]. We then concentrate the computation on the most interesting parts of the state-space, neglecting highly unlikely state-transitions. The rest of the paper presents a model-free algorithm for learning finite-state controllers of a given size. It can be used in a simulated experience protocol, as well as for learning in a direct interaction with the real environment or process.

[4] In the most general definition of finite-state automata, the next node depends not only on the previous node and observation, but also on the last action (this is the case, for instance, of graphs representing external-memory policies). The algorithm presented here can easily be generalized to this framework.

## 3 STOCHASTIC GRADIENT DESCENT IN GENERAL FINITE POLICY GRAPHS

Baird and Moore's VAPS algorithm [3] learns a reactive policy through trial-based interaction with the process to be optimized. It is based on performing stochastic gradient descent of a general error measure, and hence it can be tuned to converge to a local optimum of this error measure with probability 1. The formalism proposed encompasses any kind of error ranging from the classical Belman-residual often used in Markovian environments (i.e., TD(0)), to the TD(1) error that uses the sum of all the rewards received during the trial [29, 30]. This is the origin of the name of the algorithm: value (TD(0)) and policy (TD(1)) search. Others possible errors include those used in SARSA and in advantage learning.

Despite this robustness to the type of error used, VAPS is limited because it learns only memoryless policies. Hence it will not be effective in many partially observable environments. In this paper, we extend it so that the structure of the policy graph does not have to be completely fixed in advance, as is the case with RPs. More precisely, our algorithm can learn a general finite policy graph of a given size, possibly with simple structural constraints. We now develop the formalism of VAPS in general finite policy graphs. The presentation directly follows that of Baird and Moore [3].

### 3.1 ERROR FUNCTIONS

First we assume that the problem is a goal-achievement task, i.e., that there exists an absorbing goal-state that the system must reach as fast as possible. We also assume that the goal-state is associated with a unique observation $o_G$ that no other state produces (the system always knows with certainty when it has reached its goal). Then we can write our high-level optimality criterion as an expectation over trajectories:

$$B_\mu = \sum_{T=0}^{\infty} \sum_{\tilde{s} \in \tilde{S}_T} \Pr(\tilde{s} \mid \pi^0, \mu) \varepsilon(\tilde{s}) \ , \qquad (2)$$

where $\tilde{S}_T$ is the set of all experience sequences that terminate at time $T$, i.e.,

$$\tilde{s} = \langle o^0, n^0, a^0, r^0, ..., o^t, n^t, a^t, r^t, ..., o^T, n^T, a^T, r^T \rangle \ ,$$

$$o^T = o_G \ ,$$

and $\varepsilon(\tilde{s})$ represents the total error associated with the sequence $\tilde{s}$. [5] The total error $\varepsilon$ must be additively separable, so that

$$\varepsilon(\tilde{s}) = \sum_{t=0}^{T} e(T(\tilde{s}, t)) \ , \qquad \text{for all } \tilde{s} \in \tilde{S}_T,$$

where $e(s)$ is an instantaneous error function associated with each (finite) *sequence prefix*

$$s = \langle o^0, n^0, a^0, r^0, ..., o^t, n^t, a^t, r^t \rangle$$

($o^t$ being any observation, not necessarily the goal-observation $o_G$), and $T(\tilde{s}, t)$ represents the sequence $\tilde{s}$ truncated after time $t$. We will denote by $S_t$ the set of all sequence prefixes of length $t$ ($\tilde{S}_t \subset S_t$).

There are many possibilities for defining the immediate error $e$, including the squared Bellman residual, the error used in SARSA and the error of advantage learning (see [2, 3] for details). These three definitions make complete sense in Markovian environments only. However, they can be used for POMDPs in an approximate approach (for instance, we can use the error of SARSA to learn RPs in POMDPs). The algorithm still finds a local optimum of the error, but nothing guarantees that it will correspond to an optimal policy. The immediate error

$$e_{\text{policy}}(s) = -\gamma^t r^t \qquad \text{for all } s \in S_t,$$

induces a TD(1) search adapted to non-Markovian environment. Notably, if we use this error, then the two criteria of optimality of a policy, equation (1) and equation (2), are equal (with opposite signs however). Therefore, it will be rational to try to minimize $B_\mu$ with $e$ equal to $e_{\text{policy}}$.

### 3.2 STOCHASTIC GRADIENT DESCENT

In a general framework, $\psi$ and $\eta$ are represented as parametric functions with weights $\{w_k\}$. The objective function $B_\mu$ can be re-written as

$$B_\mu = \sum_{t=0}^{\infty} \sum_{s \in S_t} \Pr(s \mid \pi^0, \mu) e(s) \ .$$

Hence we have

$$\frac{\partial}{\partial w_k} B_\mu = \sum_{t=0}^{\infty} \sum_{s \in S_t} \left[ \Pr(s \mid \pi^0, \mu) \frac{\partial}{\partial w_k} e(s) + e(s) \frac{\partial}{\partial w_k} \Pr(s \mid \pi^0, \mu) \right] \ ,$$

[5] Note that if we are learning an RP as in the initial VAPS algorithm, then $n^t$ is completely detennined by $o^t$ and thus can be omitted in sequence $s$.

for each weight $w_k$. The partial derivative of $e$ is in general easy to calculate. In the case of $e_{\text{policy}}$ it is always 0. The only difficulty is then to differentiate $\Pr(s \mid \pi^0, \mu)$. For all $s \in S_t$ we have:

$$\Pr(s \mid \pi^0, \mu) = \prod_{j=0}^{t} \Pr(o^j \mid \pi^0, T(s, j-1))$$
$$\eta(n^{j-1}, o^j, n^j)\psi(n^j, a^j)\Pr(r^j \mid \pi^0, T(s, j-1), o^j, a^j),$$

with the conventions $T(s, -1) = \emptyset$ and $\eta(n^{-1}, o^0, n^0) = \eta^0(o^0, n^0)$.

If $\psi(n,a) > 0$ for all $(n,a)$, $\eta(n,o,n') > 0$ for all $(n,o,n')$ and $\eta^0(o,n) > 0$ for all $(o,n)$ then it can be shown that [6]

$$\frac{\partial}{\partial w_k} B_\mu = \sum_{t=0}^{\infty} \sum_{s \in S_t} \Pr(s \mid \pi^0, \mu) \left[ \frac{\partial}{\partial w_k} e(s) + e(s) \sum_{j=0}^{t} \frac{\partial}{\partial w_k} \ln \psi(n^j, a^j) + e(s) \sum_{j=0}^{t} \frac{\partial}{\partial w_k} \ln \eta(n^{j-1}, o^j, n^j) \right] . \quad (3)$$

Therefore, stochastic gradient descent of the error can be performed by repeating several trials of interaction with the process. Each experienced trial of length $T$ provides one sample of $s \in S_t$ for each $t \leq T$, which is used to estimate the expectation over $s$ in the above equation. Of course these samples are not independent, but it does not introduce any bias since we sum the different estimates. During each trial, the weights are kept constant and the approximate gradients of the error at each time $t$

$$\frac{\partial}{\partial w_k} e(s) + e(s) \sum_{j=1}^{t} \left( \frac{\partial}{\partial w_k} \ln \psi(n^{j-1}, a^{j-1}) + \frac{\partial}{\partial w_k} \ln \eta(n^{j-1}, o^j, n^j) \right)$$

are accumulated. Weights are updated at the end of each trial, using the sum of these immediate gradients. An incremental implementation of the algorithm can be obtained by using, at every step $t$, the following update rules:

$$\Delta T_k^\psi(t) = \frac{\partial}{\partial w_k} \ln \psi(n^{t-1}, a^{t-1})$$
$$\Delta T_k^\eta(t) = \frac{\partial}{\partial w_k} \ln \eta(n^{t-1}, o^t, n^t)$$
$$\Delta w_k(t) = -\alpha \left[ \frac{\partial}{\partial w_k} e(s_t) + e(s_t)(T_k^\psi(t) + T_k^\eta(t)) \right] ,$$

where $s_t = \langle o^0, n^0, a^0, r^0, ..., o^t, , n^t, a^t, r^t \rangle$ represents the experience prefix at time $t$, $\alpha$ is the step-size parameter (or learning rate), and $T_k^\psi$ and $T_k^\eta$ are the trace sassociated with weight $k$ in the representation of $\psi$ and $\eta$, respectively. The complete policy-update performed at the end of the trial is then given by

$$\Delta w_k = \sum_{t=0}^{T} \Delta w_k(t) \ ,$$

8 where $T$ is the length of the trial. Note that the traces $T_k^\psi(t)$ and $T_k^\eta(t)$ are independent of the immediate error $e$ used. They only depend on the way the policy-graph parameters vary with the weight $w_k$, i.e., on the representation chosen for these parameters. The main novelty of our algorithm (compared to the original VAPS) is the use of a second trace ($T^\eta$), which is analogous to the original trace ($T^\psi$), but summarizes the node-transition executed during the trial instead of the action-choices.

### 3.3 EXAMPLES

If we use look-up tables to store the parameters of the policy graphs, then there is one weight, denoted $\psi_{n,a}$, for each possible $(n,a)$, one weight $\eta_{n,o,n'}$ for each possible $(n,o,n')$, and one weight $\eta^0_{o,n}$ for each $(o,n)$, such that $\psi(n,a) = \psi_{n,a}$, $\eta(n,o,n') = \eta_{n,o,n'}$ and $\eta^0(o,n) = \eta^0_{o,n}$. Suppose also that we are using the immediate error $e_{\text{policy}}$, i.e., we are performing a TD(1) search. Then the contribution to the update of each weight at each time-step in the sequence can be expressed as:

$$\Delta\psi_{n,a}(t) = -\alpha\gamma^t r^t \frac{N_{n,a}(t)}{\psi_{n,a}} \ ,$$
$$\Delta\eta_{n,o,n'}(t) = -\alpha\gamma^t r^t \frac{N_{n,o,n'}(t)}{\eta_{n,o,n'}} \ ,$$
$$\Delta\eta^0_{o,n}(t) = -\alpha\gamma^t r^t \frac{N^0_{o,n}}{\eta^0_{o,n}} \ ,$$

where $N_{n,a}(t)$ is the number of times that action $a$ has been executed in node $n$ at time $t$, $N_{n,o,n'}(t)$ is the number of times that we moved from node $n$ to node $n'$ after observation $o$ between time 0 and $t$, and $N^0(o,n) = 1$ if $o^0 = o$ and $n^0 = n$, and $N^0(o,n) = 0$ otherwise.

Despite its simplicity, the look-up table representation has several drawbacks. First, the weights $w_k$ represent probabilities, and thus they are subject to constraints. As a matter of fact, nothing guarantees that the probabilities will still belong to [0, 1] and sum to 1 if we apply the update rule described above. A classical solution to this problem involves projecting the gradient on the simplex before applying it. However it does not eradicate the second drawback of the look-up table representation, i.e., there is still no guarantee that $\psi(n,a) > 0$ and $\eta(n,o,n') > 0$ for all $n$, $a$, $o$, $n'$. Hence, the derivative of $B_\mu$ may not be equal to equation

[6] If this condition on $\psi$ and $\eta$ is not satisfied, then there exist zero probability trajectories that have a non-zero contribution to the gradient [21, 15].

(3) in all the points of the policy-graph space. Studying how to express the gradient in such cases falls beyond the scope of this paper (see [21, 15]).

In our experiments, we use the soft-max function (or Boltzmann law) to represent the parameters of the graphs. In this case, the weights $w_k$ are "Q-values" $Q^{\psi}(n,a)$, $Q^{\eta}(n,o,n')$ and $Q^{\eta^0}(o,n)$ such that

$$\psi(n,a) = \frac{e^{Q^{\psi}(n,a)/\theta}}{\sum_{a' \in A} e^{Q^{\psi}(n,a')/\theta}} ,$$

$$\eta(n,o,n') = \frac{e^{Q^{\eta}(n,o,n')/\theta}}{\sum_{n'' \in N} e^{Q^{\eta}(n,o,n'')/\theta}} ,$$

$$\eta^0(o,n) = \frac{e^{Q^{\eta^0}(o,n)/\theta}}{\sum_{n' \in N} e^{Q^{\eta^0}(o,n')/\theta}} ,$$

where $\theta$ is a temperature parameter. Although it complicates the calculation of the gradient slightly, this representation avoids both problems of look-up tables: the Q-values can take any real values, and the induced policy never gives probability 0 to any choice. The use of the Boltzmann law may strongly modify the shape of the error function with respect to the weights $w_k$. Hence, it influences the performance of gradient algorithms such as VAPS. It is difficult to say *a priori* if its influence will be beneficial or negative, for a given problem.

### 3.4 VARIANTS AND REMARKS

It is straightforward to extend the algorithm so that it handles simple constraints on the policy graph. If we constraint the graph to represent a RP, then the algorithm is equivalent to Baird and Moore's original VAPS. Consider as another example the finite external-memory architecture used by Peshkin et al. [23]. There are two ways to model this architecture: either we augment the POMDP state-, observation- and action-spaces but still use a RP, or we leave the POMDP unchanged and use a more complex policy graph than a simple RP (this graph contains $2^L|O|$ nodes, where $L$ is the number of external memory bits). In the first case, the probability of changing the content of the memory is represented in $\psi$, in the second case it is represented in $\eta$. Our results are coherent in the sense that, as the update rule uses $\psi$ and $\eta$ in a completely similar way, the algorithm will be the same whatever the interpretation chosen. Another possibility is to learn finite RP-sequences such as in HQL, either using $e_{\text{policy}}$, or defining a new error function $e_{\text{HQL}}$ based on the HQ-values of the algorithm. In the first case, we will converge to an RP-sequence which is locally optimal in the sense of the expected total reward (1). In the second, we will find a local minimum of the error, but it may not correspond to a policy that maximizes (even locally) the expected discounted reward.

Another question is how to treat discounted problems where there is no goal state, and, therefore, no natural notion of trial (the so-called maintainance tasks). One possibility for dealing with discounted maintainance task is the following: at each time step we execute an independent random drawing to determine if we terminate the trial. We set the probability to end the trial to be constant and equal to $(1-\gamma)$ and we do not discount the rewards received during the trial, i.e., we use

$$e'_{\text{policy}} \stackrel{\text{def}}{=} -r^t .$$

Then the policy (graph) that maximizes $B_\mu$ is also an optimal policy graph in the usual sense (equation (1)). This trick allows TD(1) learning of maintainance tasks, but it is not adapted to other kinds of immediate error $e$. Baird and Moore argue that VAPS can be adapted to discounted POMDPs whatever the immediate error, but it is not clear to us how to do this without introducing a bias in the estimates.

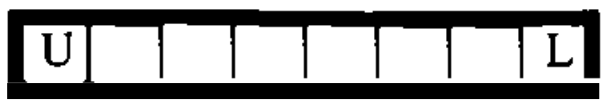

Figure 2: The load/unload problem with 8 locations: the agent starts in the "Unload" location (U) and receives a reward each time it returns to this place after passing through the "Load" location (L). The problem is partially observable because the agent cannot distinguish the different locations in between Load and Unload, and because it cannot perceive if it is loaded or not ($|S| = 14$, $|O| = 3$ and $|A| = 2$).

## 4 NUMERICAL SIMULATIONS

In this section, we present the results of two experiments. The first aims at comparing an exact gradient algorithm [22] with the stochastic gradient approach of VAPS. The second shows that our algorithm can solve a moderately-difficult real-world problem. In all the experiments, we used the immediate error $e_{\text{policy}}$ and we initialized the policy graph with uniform distributions.

### 4.1 COMPARISON WITH EXACT GRADIENT DESCENT

A model-free learning algorithm such as VAPS may be used to learn a policy when we do not know all the parameters of the POMDP in advance. As explained in section 2.3, it is also useful when the problem is perfectly known in advance: the protocol of simulated experience allows optimizing huge problems with sparse structure, by sampling only the probable trajectories, instead of considering all trajectories. It is interesting to look at the conditions under which VAPS would be expected to outperform the exact algorithm.

First, it is to be noted that the exact gradient calculation is very sensitive to the size of the state space of the POMDP: each step of the computation has complexity at least in $O(|S|)$. The influence of the size of $S$ on VAPS is less clear: the complexity of updating the weight is independent of $|S|$, however, a bigger state-space would require (and induce) longer experience trials. In practice, it has been very easy for us to build a problem with many states and few observations where VAPS completely outperforms the exact gradient (in terms of real computing time). Therefore, the first rule is that VAPS scales up much better than exact gradient to problems with big state-spaces. This is not surprising since handling big state-spaces was precisely the original motivation of this work.

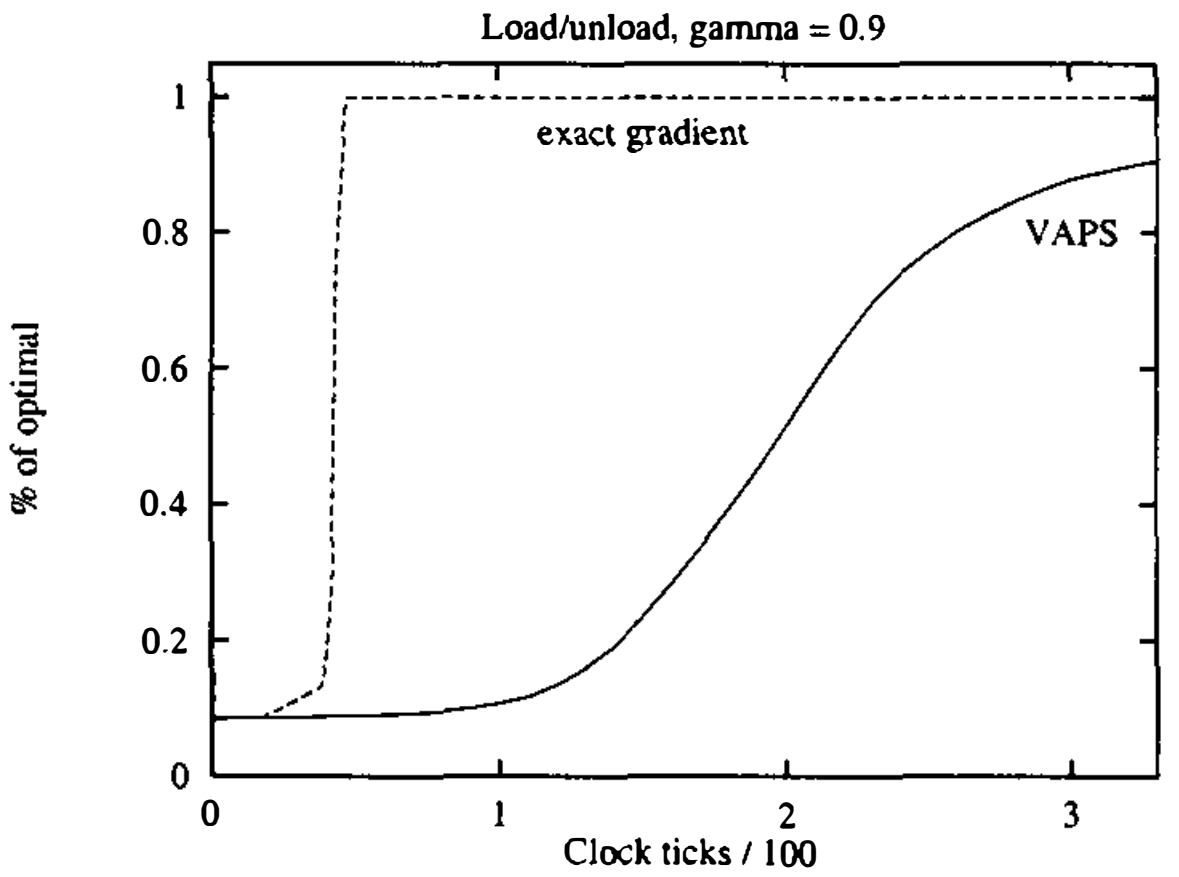

Figure 3: Learning curves of VAPS and exact gradient descent on the load/unload problem, with $\gamma$ = 0.9: $\theta = 1$, $\alpha$ is chosen at its optimal value (for each algorithm), and the results of stochastic gradient is averaged over 30 experiences.

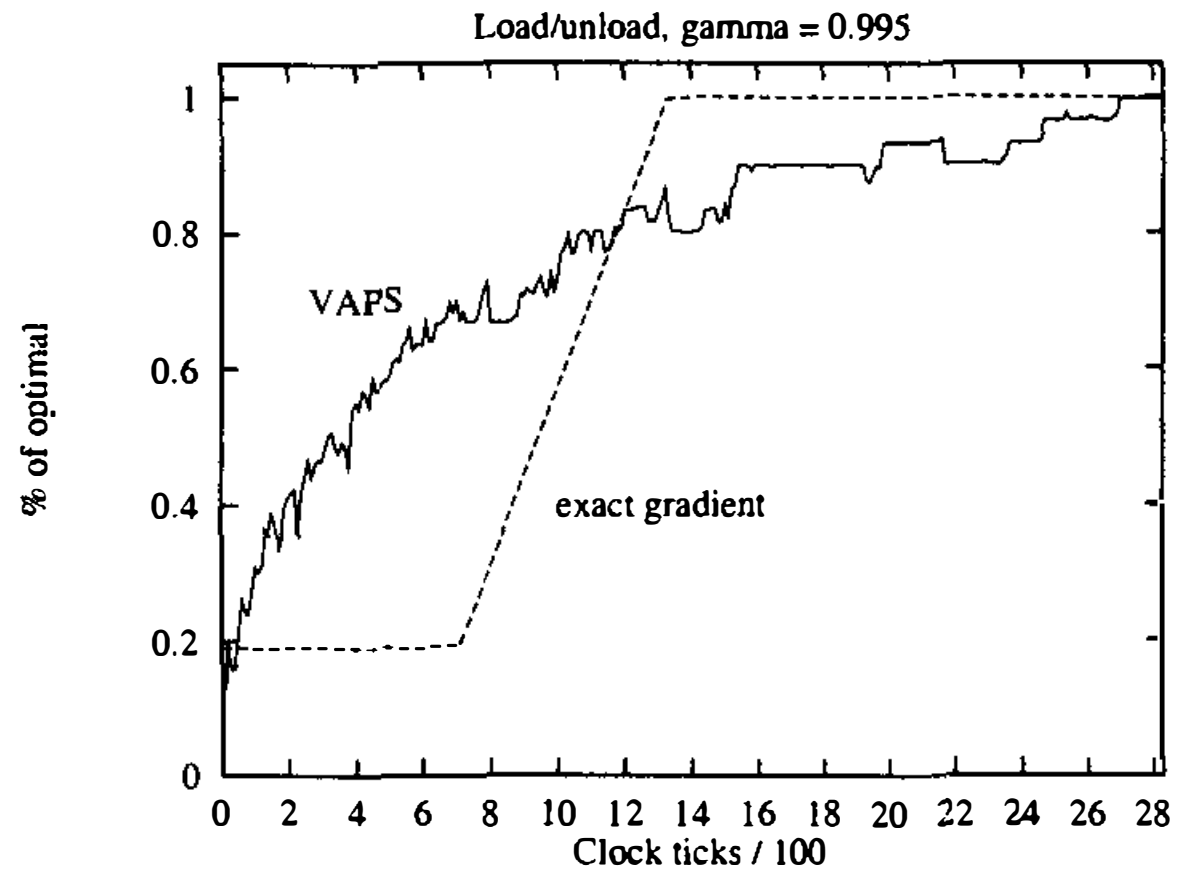

Figure 4: Same as figure 3, but $\gamma = 0.995$.

The second important variable in our comparison is the discount factor $\gamma$. In general, a bigger $\gamma$ helps (both exact and stochastic) gradient based algorithms because it increases the value function and thus makes the gradient steepest. However, $\gamma$ may have many other (contradictory) effects on the algorithms. In the case of VAPS, as the trials are ended with probability $1-\gamma$, bigger $\gamma$ will make longer and hence more instructive trials. On another hand, the exact gradient calculation requires solving several Belman equations in the cross-product MDP (cf. section 2.3, [22]). This is done by successive approximation (or value iteration), which is very sensitive to $\gamma$. The bigger the $\gamma$, the more iterations needed to reach a given accuracy. There are then two opposite tendencies in both algorithms: increasing $\gamma$ could accelerate them as well as slow them down.

To clarify this point, we ran the exact gradient algorithm and VAPS on the simple load/unload problem presented in figure 2 (with 5 locations). The number of nodes of the graph is fixed to 2, which is the optimal number for this problem. We tried several values of $\gamma$ ranging from 0.9 to 0.995, and plotted the learning curves produced by both algorithms. These learning curves represent the evolution of the performance of the current policy (expressed as a percentage of the optimal performance), as a function of the real-time spent learning (expressed in computer time-ticks). Although very simple, load/unload provides an illustration of the mechanism depicted above. With $\gamma = 0.9$ the exact gradient clearly outperforms VAPS (cf. figure 3). When $\gamma$ increases, the difference between the algorithm vanishes. At $\gamma = 0.995$ (figure 4), the two techniques are roughly equivalent. Beyond this point, VAPS dominates exact gradient descent. The conclusion of this experiments is that the execution time of both exact and stochastic gradient descent do increase with $\gamma$, but this increase is more dramatic in the case of exact gradient.

As a conclusion, stochastic gradient descent can outperform exact gradient in problems with large state space and large discount factor.

### 4.2 EXPERIMENTS WITH POLE BALANCING

We ran a number of experiments with the pole balancing problem [30]. This famous problem is known to be solved by an RP, if the observation at each time-step is composed of four elements: the cart position $x$ and speed $\dot{x}$, and the pole angle $\phi$ and angular speed $\dot{\phi}$. To measure the difficulty of the task and the performance of our algorithm, we used two different settings: a completely observable one where the four relevant variables $x$, $\dot{x}$, $\theta$ and $\dot{\theta}$ can be seen by the algorithms at each time-step, and a partially observable setting where both $\dot{x}$ and $\dot{\theta}$ are always hidden.

We ran three different algorithms in both settings: SARSA, Baird and Moore's original VAPS (learning an RP) and our extension of VAPS allowing to learn policy graphs, varying the number of nodes of the graph. SARSA and the original VAPS can be expected to succeed in the completely observable setting, and to fail in the partially observable one where there is no reactive policy that performs the task.

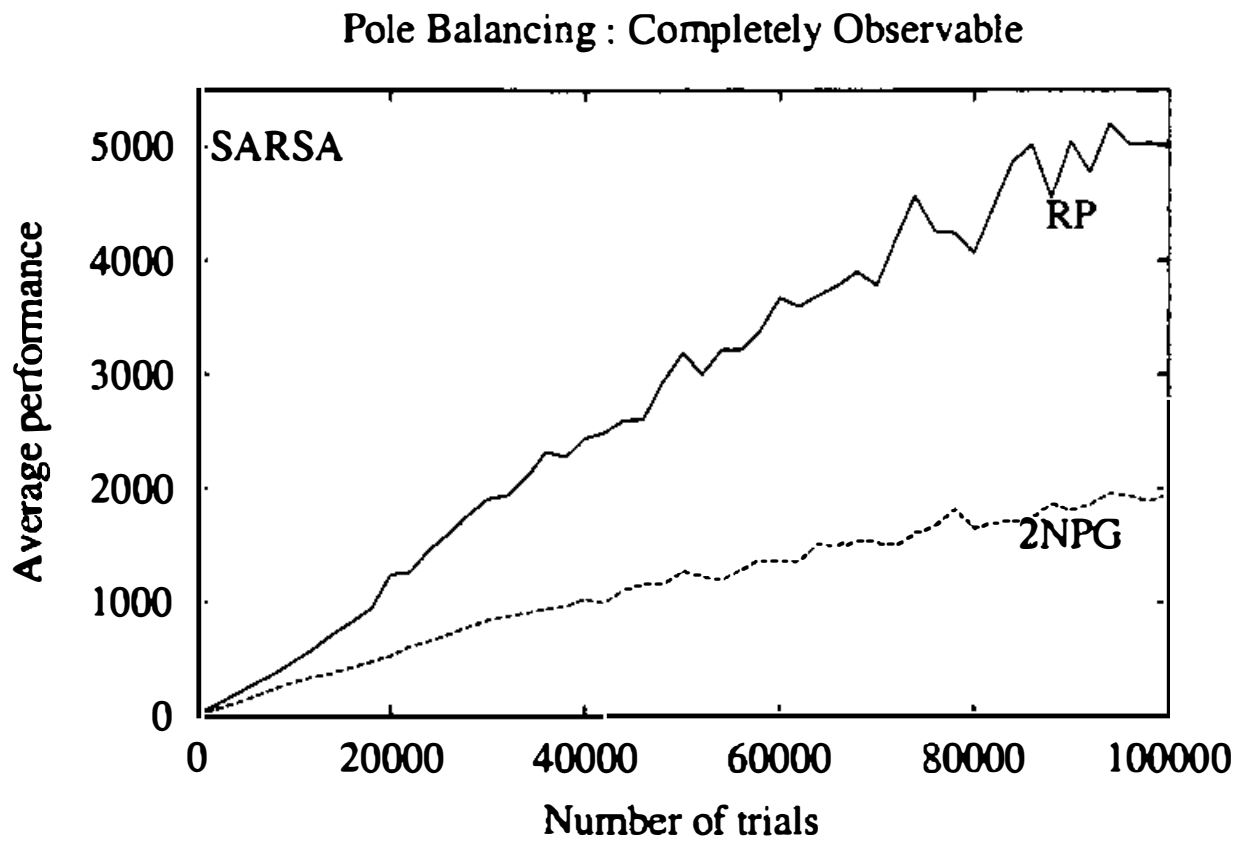

Figure 5: Learning curves obtained with the completely observable pole-balancing problem. "RP" stands for the original VAPS algorithm, as proposed by Baird and Moore; "2NPG" represents our extension of VAPSused with $N = 2$.

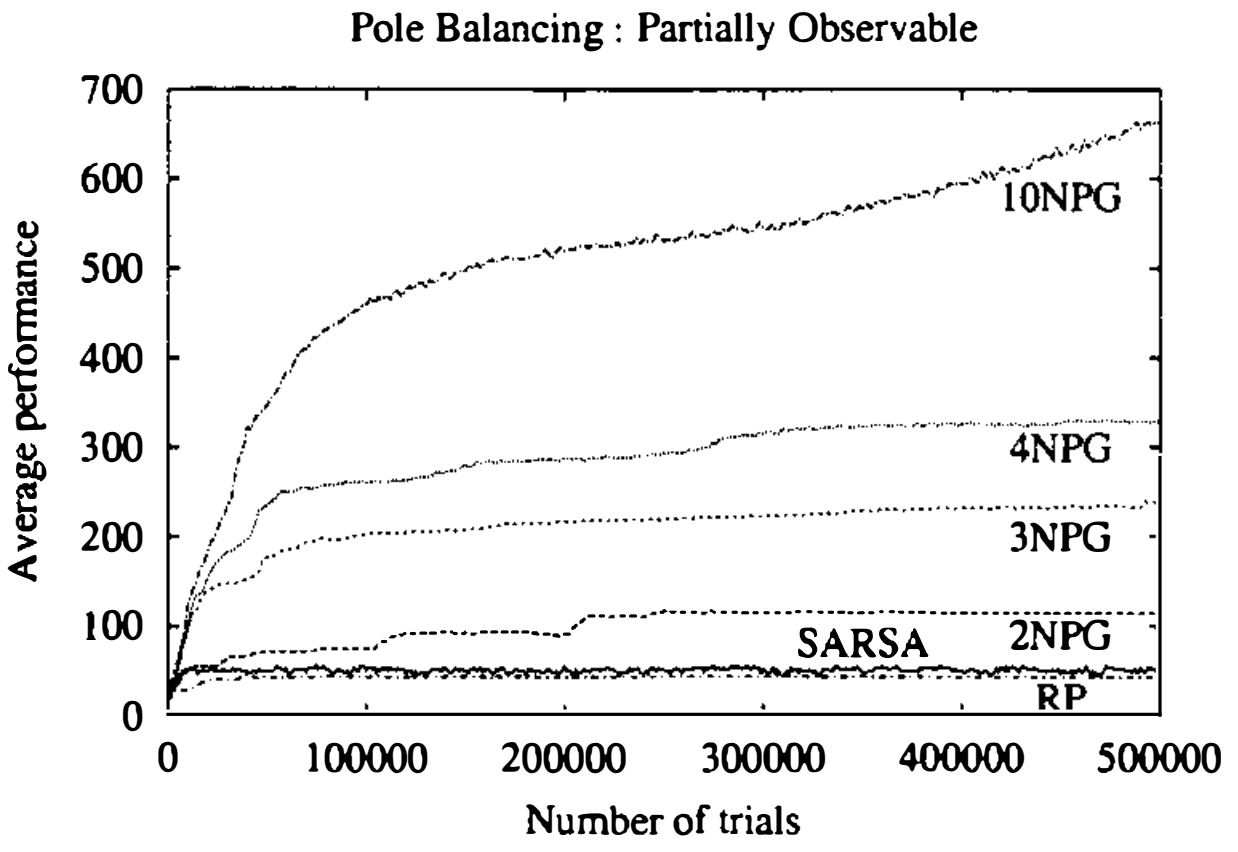

Figure 6: Learning curves obtained with the partially observable pole-balancing problem. "RP" stands for the original VAPS algorithm, as proposed by Baird and Moore; "2NPG", "3NPG" , "4NPG" and "10NPG"represent our extension of VAPSused 2, 3, 4 and 10 nodes respectively.

These two algorithms differ radically. On one hand we use VAPS with the immediate error $e_{\text{policy}}$ which makes it equivalent to TD(1), Baird and Moore would call this a pure policy search. On the other hand SARSA is basically a value-search in the line of TD(0). Our algorithm can be expected to succeed in both settings, provided that we use a sufficiently large policy graph, and that algorithm does not get stuck on a local optimum. Two nodes should be enough in the completely observable setting, since every reactive policy using only two actions (as it is the case here) can be represented by a two-node policy graph. In the partially observable framework, more nodes must be added to allow the algorithm memorize past observations.

In all experiments the discount factor $\gamma$ was set to .99 and increased gradually as learning progressed. The learning rate $\alpha$ was optimized independently for each algorithm. The performance of the algorithm was measured by fixing the policy and executing 200 trials, measuring the length of each trial in terms of control decisions, and averaging these measures. The value intervals of cart position and pole position were partitioned into 6 and 3 unequal parts (smaller size of partition towards the center) in the completely observable setting, and into 8 and 6 parts in the partially observable setting, correspondingly. We were making decisions at the rate of 50 Hz, meaning, for example, that the actual physical time of learning to balance a pole for 500 sequential ticks corresponds to 10 seconds of balancing. Other parameters of the cart and pole balancing problem were taken as described in the supplementary WWW page for [30].

Figure 5 presents the learning curves obtained in the completely observable framework. The horizontal axis represents the number of trials, which corresponds to the number of times we have dropped the pole. The vertical axis represents the performance of the algorithm, measured as explained above. We see that:

- SARSA learns much faster than the original VAPS, showing that value search is much more efficient that policy search for this control problem,
- our extension of VAPS with 2-node policy graph learns slower than the original VAPS. This phenomenon can be explained by the fact that the space of 2-nodes policy graphs is bigger than the space of RPs.

Figure 6 presents the results obtained in the partially observable framework. These results confirm our expectation that algorithms limited to reactive policy will fail. In contrast, our algorithm increases its performance gradually, showing that it is able to compensate the lack of observability. The more nodes are given to the algorithm, the better it performs. It is also striking to see that the performance of the algorithm seems to improve by steps, which makes difficult to predict where learning will stop. Because of limited time, we could not continue the experiments beyond 500000 iterations so that we do not know if the performance would continue to increase until the system may balance infinitely long. We are currently running this experiment and the results will be shown in a forthcoming technical report [15]. The most significant current result is that we can learn the structure of the policy graph that extracts some useful information contained in the string of past observations, to compensate, at least partially, for the lack of observability. Pole balancing is a widely accepted benchmark problem for dynamic system control and to the best of our knowledge it has not been learned with partial information.

## 5 CONCLUSION

We have derived an extension of a general algorithm that enables it to learn policies using a memory. The basic principle of this algorithm is to perform stochastic gradient descent on finite-state controller parameters, which guarantees local optimality of the solution produced. Moreover, we have led an experimental study of this approach, and compared it to classic (non-adaptive) algorithms in terms of execution time and learning speed. At last, we showed that our algorithm can solve a difficult problem such as pole-balancing without having access to all the information usually required to solve it. Therefore, it is able to find the structure of the policy graph that extracts the useful information contained in the sequence of past observations to compensate for the lack of observability at each time-step. We believe that this constitutes a significant achievement and proves that our algorithm can be efficient in some real-world problems.